\documentclass[lettersize,journal]{IEEEtran}
\usepackage{etoolbox}
\makeatletter
\@ifundefined{color@begingroup}%
{\let\color@begingroup\relax
\let\color@endgroup\relax}{}%
\def\fix@ieeecolor@hbox#1{%
\hbox{\color@begingroup#1\color@endgroup}}
\patchcmd\@makecaption{\hbox}{\fix@ieeecolor@hbox}{}{\FAILED}
\patchcmd\@makecaption{\hbox}{\fix@ieeecolor@hbox}{}{\FAILED}
\usepackage{amsmath,amsfonts}
\usepackage{algorithmic}
\usepackage{algorithm}
\usepackage{amsmath}
\usepackage{subfigure}  %插入多图时用子图显示的宏包
\usepackage{array}
\usepackage[table,xcdraw]{xcolor}
\usepackage{multirow}
\usepackage[defaultcolor=red]{changes}  %使用changes宏包
\definechangesauthor[name={Author}, color=red]{A} %修订作者

\usepackage[colorlinks, linkcolor=blue, citecolor=blue, urlcolor=blue, filecolor=blue]{hyperref}
\usepackage{mathrsfs}
\usepackage{textcomp}
\usepackage{stfloats}
\usepackage{url}
\usepackage{verbatim}
\usepackage{graphicx}
\usepackage{cite}
\usepackage{orcidlink}
\usepackage{cleveref}
\usepackage{subcaption}
\usepackage{float}
\usepackage{bbding}
\usepackage{amssymb}
\usepackage{setspace}
\usepackage{tikz}
\usepackage{setspace} % 控制行距
\usetikzlibrary{tikzmark,calc}
\usepackage{amssymb} % for \checkmark
\usepackage{caption}
\begin{document}

\title{RSNet: A Light Framework for The Detection of SAR Ship Detection}

\author{Hongyu Chen\textsuperscript{\large\orcidlink{0009-0004-2934-3135}}, 
Chengcheng Chen\textsuperscript{\large\orcidlink{0000-0002-6014-5135}}, \IEEEmembership{Graduate Student Member, IEEE}, 
Fei Wang\textsuperscript{\large\orcidlink{0009-0005-3061-0496}},\\
Yuhu Shi\textsuperscript{\large\orcidlink{0000-0002-4009-2849}}, and Weiming Zeng\textsuperscript{\large\orcidlink{0000-0002-9035-8078}}, \IEEEmembership{Senior Member, IEEE}

\thanks{This work was supported by the National Natural Science Foundation of China (grant nos. 31870979), in part by the 2023 Graduate Top Innovative Talents Training Program at Shanghai Maritime University under Grant 2023YBR013. (Corresponding author: Weiming Zeng)}
\thanks{ Hongyu Chen, Chengcheng Chen, Fei Wang, Yuhu Shi, Weiming Zeng are with the Laboratory of Digital Image and Intelligent Computation, Shanghai Maritime University, Shanghai 201306, China (e-mail: hongychen676@gmail.com, shmtu\_ccc@163.com, shine\_wxf@163.com, syhustb2011@163.com, zengwm86@163.com).}

 % <-this % stops a space

%\thanks{This paper was produced by the IEEE Publication Technology Group. They are in Piscataway, NJ.}% <-this % stops a space
\thanks{Manuscript received XX, XXXX; revised XXXX XX, XXXX.}}

\markboth{Journal of \LaTeX\ Class Files,~Vol.~XX, No.~X, Oct~2024}%
{Shell \MakeLowercase{\textit{et al.}}: A Sample Article Using IEEEtran.cls for IEEE Journals}


\maketitle

\begin{abstract}
Recent advancements in synthetic aperture radar (SAR) ship detection using deep learning have significantly improved accuracy and speed, yet effectively detecting small objects in complex backgrounds with fewer parameters remains a challenge. This letter introduces RSNet, a lightweight framework constructed to enhance ship detection in SAR imagery. To ensure accuracy with fewer parameters, we proposed Waveletpool-ContextGuided (WCG) as its backbone, guiding global context understanding through multi-scale wavelet features for effective detection in complex scenes. Additionally, Waveletpool-StarFusion (WSF) is introduced as the neck, employing a residual wavelet element-wise multiplication structure to achieve higher dimensional nonlinear features without increasing network width. The Lightweight-Shared (LS) module is designed as detect components to achieve efficient detection through lightweight shared convolutional structure and multi-format compatibility. Experiments on the SAR Ship Detection Dataset (SSDD) and High-Resolution SAR Image Dataset (HRSID) demonstrate that RSNet achieves a strong balance between lightweight design and detection performance, surpassing many state-of-the-art detectors, reaching 72.5\% and 67.6\% in \textbf{\(\mathbf{mAP_{.50:.95}}\) }respectively with 1.49M parameters. Our code will be released soon.
%RSNet 以 Waveletpool-ContextGuided (WCG) 为骨干，通过小波背景引导策略以达到更少的参数提高准确性，并以 Waveletpool-StarFusion (WSF) 为头，有效减少参数。

\end{abstract}


\section{Introduction}
\IEEEPARstart{S}{ynthetic} Aperture Radar (SAR) imagery is a vital tool in remote sensing \cite{moreira2013tutorial}, offering high-resolution images under all weather and lighting conditions. By using microwave signals that can penetrate atmospheric obstacles like clouds and rain, SAR provides detailed insights into terrain \cite{dobson1995land}, vegetation \cite{bao2023vegetation}, and man-made structures \cite{connetable2024corrections}. Its applications span geological exploration \cite{baraha2023synthetic}, disaster monitoring \cite{karimzadeh2022deep}, agriculture \cite{hashemi2024review}, and urban planning. Notably, SAR excels in ocean monitoring \cite{asiyabi2023synthetic}, detecting subtle surface deformations, and generating digital elevation models, making it indispensable for modern earth observation.

With SAR image data scale and complexity increasing, traditional image processing methods are becoming insufficient \cite{fracastoro2021deep}. Recent advances in convolutional neural networks (CNNs) and Transformer architectures have significantly improved target detection \cite{wang2024hybrid}, \cite{du2021two}, \cite{zhang2024mgsfa}, \cite{shen2024ellk}. These deep learning methods are generally categorized into candidate region-based and regression-based approaches, with the latter excelling in real-time detection by directly regressing targets at multiple scales. For SAR object detection, one-stage methods strike a promising balance between speed and accuracy; however, due to the computational and energy limitations of airborne platforms, a lightweight model framework is essential for practical deployment.

Building on the established model framework, several notable improvements have been proposed. For example, Tian et al. \cite{tian2024faster} introduced LFer-Net, which integrates SPDConv and InceptionDWConv for high accuracy and efficiency. Zhou et al. \cite{zhou2023hrle} proposed HRLE-SARDet, leveraging hybrid representation learning to improve speed-accuracy balance. Chang et al. \cite{chang2023mlsdnet} presented MLSDNet, utilizing Adaptive Scale Distribution Attention (ASA) to enhance generalization and detection precision. Feng et al. \cite{feng2022lightweight} developed LPEDet, an anchor-free SAR ship detection model with a novel position-enhanced attention strategy. Yang et al. \cite{yang2022efficient} designed a lightweight network featuring bidirectional pyramidal structures and soft quantization to minimize accuracy loss and background interference.

While advancements have shown the viability of lightweight models for SAR detection, substantial improvement opportunities remain. SAR images, unlike optical images, are prone to speckle noise, which reduces detail resolution \cite{moreira2013tutorial}, and face increased interpretation complexity due to geometric distortions. Furthermore, the processing of grayscale SAR data lacks intuitive color information. These factors hinder the balance between accuracy and detection speed, particularly regarding the crucial metric $\mathrm{mAP_{.50:.95}}$ for optimizing detection performance.

To tackle challenges in remote sensing image detection, we present RSNet (Restar the Sky Network), an optimized network based on YOLOv8, featuring improvements to its backbone, neck, and head. The backbone incorporates the Waveletpool-ContextGuided (WCG) structure, while the head employs the Waveletpool-StarFusion (WSF) network. Additionally, a Lightweight-Shared (LS) module in the detection head reduces the model's parameter load.

The primary contributions of this article are as follows:

\begin{itemize}
    \item We developed the WCG architecture, enhancing detection accuracy through anti-aliasing and contextual integration, achieving superior performance with fewer parameters.
    
    \item The WSF network is constructed by combining wavelet upsampling pooling and element-wise multiplication through the residual structure, significantly reducing parameters while maintaining computational efficiency.
    
    \item We introduced the LS module using the group normalization strategy to further minimize the parameter burden of the detection head, thereby enhancing the overall efficiency of RSNet.
    
    \item Experiments conducted on the SSDD and the HRSID show that the proposed RSNet strikes an improved balance between lightweight design and detection performance.
\end{itemize}
\section{PROPOSED METHOD}
\begin{figure}
\small 
	\centering
	\includegraphics[width=\linewidth]{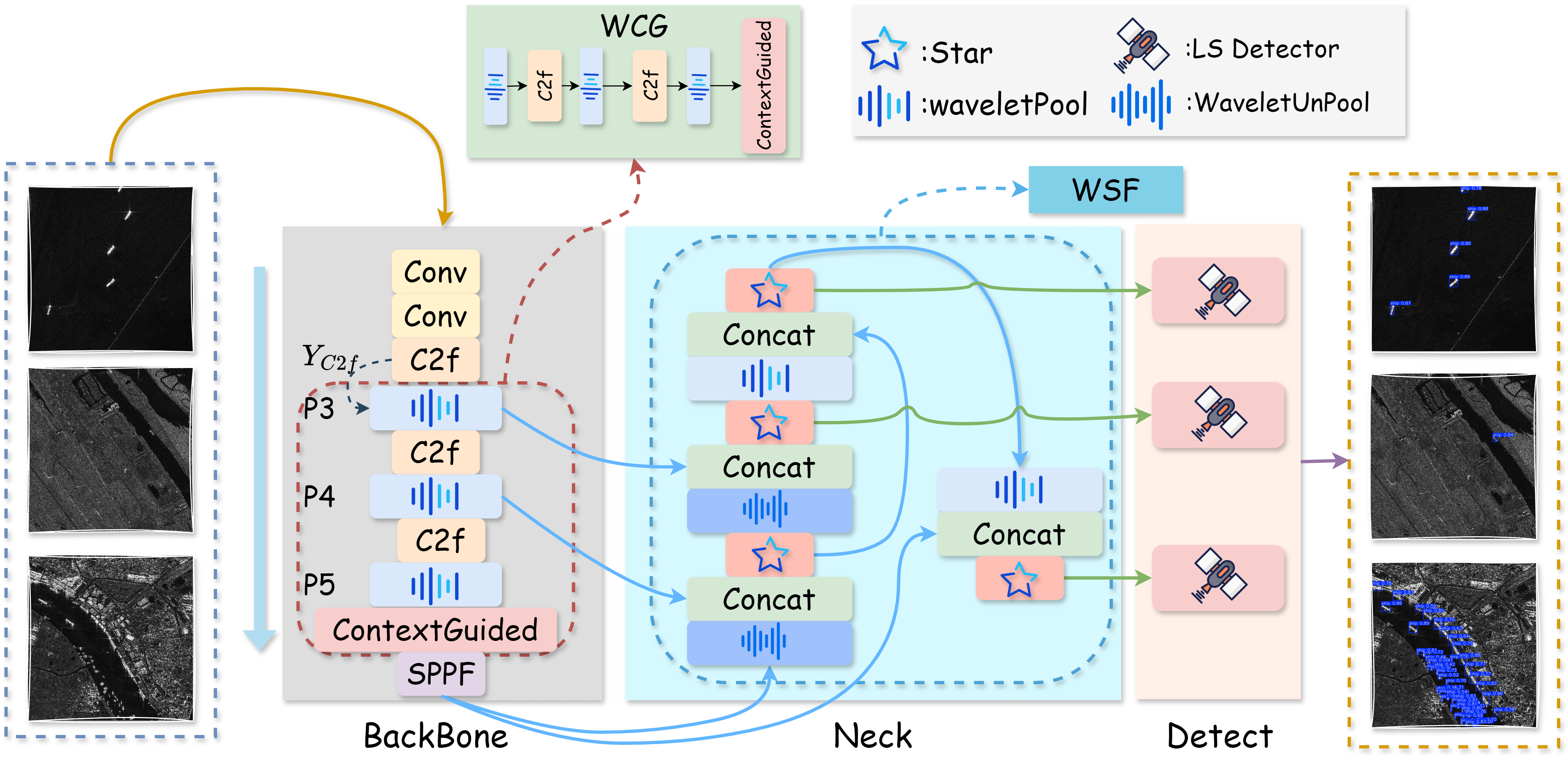}
	\caption{Overall flow of the RSNet.}
	\label{fig:fig1}
	\medskip
\end{figure}

\subsection{Overall Architecture}
This letter introduces RSNet (Restar the Sky Network), a lightweight network for ship detection in SAR images, structured into backbone, neck, and detect components as illustrated in Fig. \ref{fig:fig1}. The backbone processes the input image through the first three layers to produce feature maps, which is then enhanced by multiple scales of wavelet pooling to generate multi-scale feature maps. The ContextGuided block further enriches feature extraction by integrating contextual information. In the neck, the WSF module employs wavelet upsampling to restore spatial resolution and fuse features.  The lightweight star module combines an element-wise multiplication structure with skip connections for deeper feature extraction, enhancing small object recognition capability. Finally, the LS Detector module connects the features to output the bounding boxes, class labels, and confidence scores of the detected objects.

\subsection{Waveletpool-ContextGuided (WCG)}
In complex backgrounds, including land structures, clutter, and coastlines, current SAR ship detectors face challenges in accurately locating ships. To achieve better performance with fewer parameters, this letter proposes WCG.

The WCG network introduces a wavelet pooling mechanism that efficiently extracts multi-scale spatial features, reducing computational complexity while enhancing sensitivity to subtle details in complex backgrounds. This allows the network to focus on potential ship targets in noisy environments. Additionally, the context guided convolutional module integrates surrounding environmental information, significantly improving detection of small targets, especially in high-noise conditions. By leveraging contextual information, the model better distinguishes real ship targets from background noise, enhancing localization accuracy. Below are specific descriptions of wavelet pooling and context guided convolution.

\begin{figure*}
	\centering
	\includegraphics[width=0.95\textwidth]{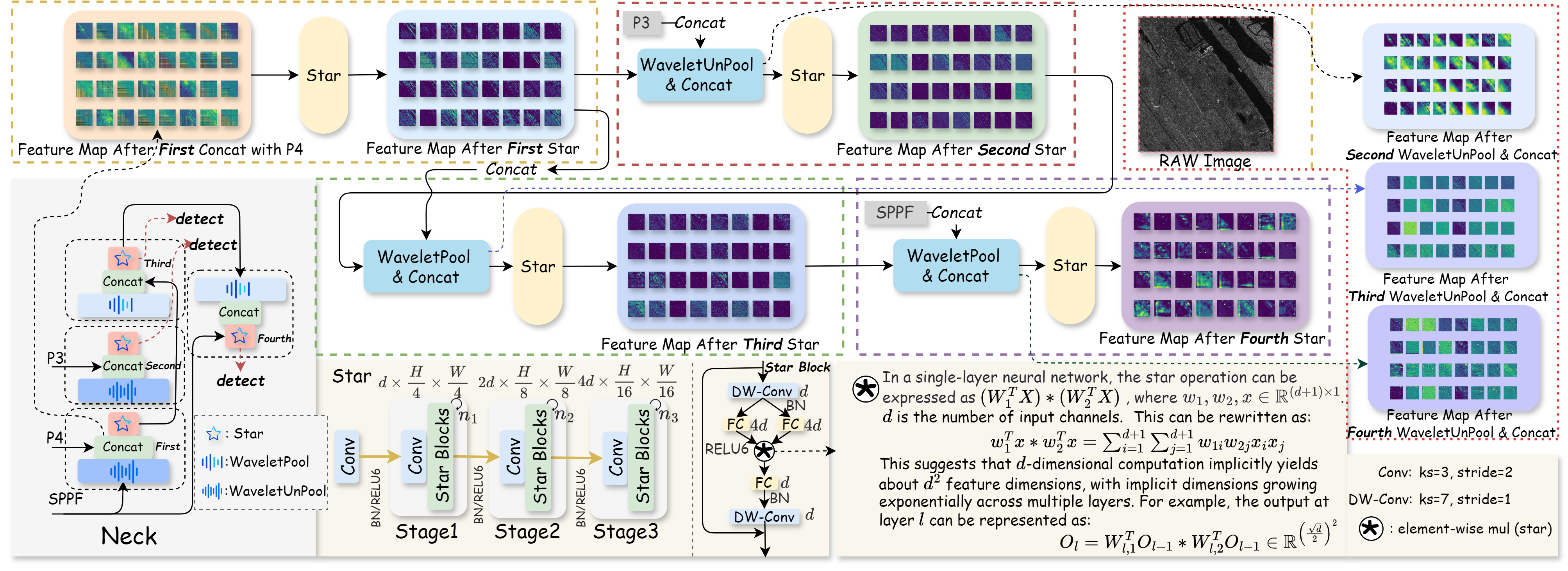}
	\caption{Flowchart of WSF (Neck of RSNet). The feature map from the backbone network is processed by a residual structure comprising four submodules: WaveletUnPool, Concat, and Star. The network structure of the Star module is also illustrated.}
	\label{fig:fig2}
	\medskip
	\small 
\end{figure*}
\begin{figure}
\small 
	\centering
	\includegraphics[width=\linewidth]{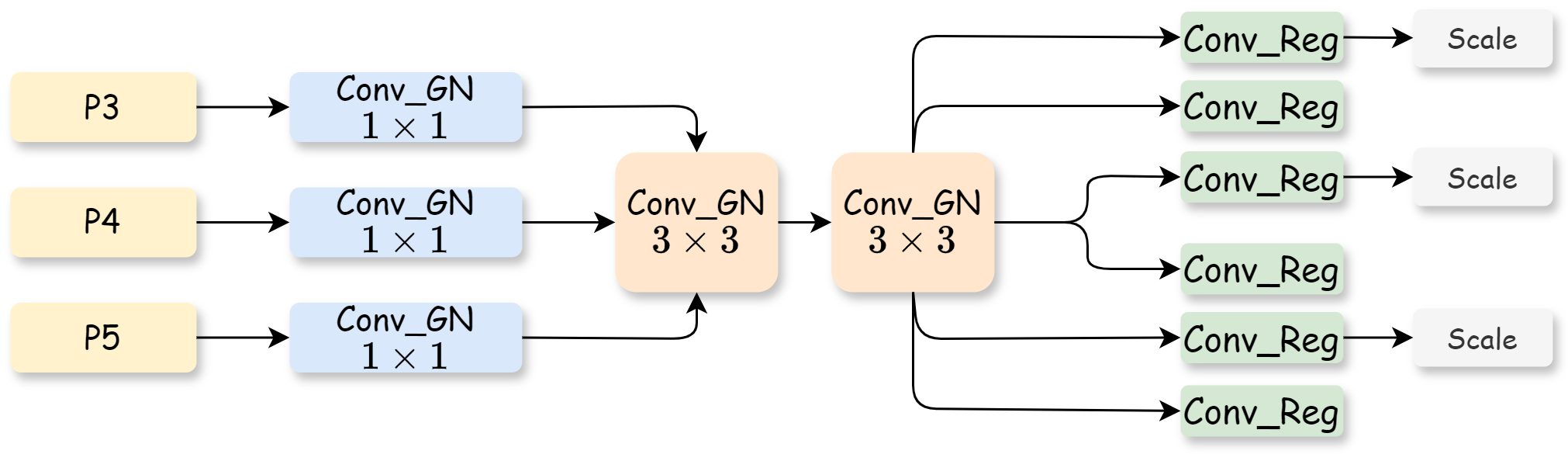}
	\caption{The structure of LS Detector.}
	\label{fig:fig3}
	\medskip
\end{figure}
\subsubsection{Wavelet Pooling}
In the WCG, the WaveletPool module plays a critical role, designed to effectively extract multi-scale features from the input feature map through wavelet transforms. The output $Y_{pool}$ of this module is defined by the following formula:
\vspace{-0.1cm}
\begin{equation}
\begin{aligned}
    Y_{{pool}}={Conv}_{2{D}}(Y_{{C2f}})\in{R}^{B\times C_{{out}}\times H\times W},
    %Y_{{pool}}={Conv}_{2{D}}(Y_{{C2f}},F,{stride}=2,{groups}=C),\\
   % Y_{{pool}}\in{R}^{B\times C_{{out}}\times H^{\prime}\times W^{\prime}},\\
\end{aligned}
\label{eq:1}
\end{equation}
where we define the output feature map from C2f as $Y_{{C2f}}$. $B$ is the batch size. $H$ and $W$ are the dimensions of the downsampled output. $C_{{out}}$ is the number of output channels of the WaveletPool block. The filter set ${F}$ used in the WaveletPool operation consists of four different convolutional kernels defined as:
\vspace{-0.4cm}
\begin{equation}
\begin{aligned}
    F=\{F_{LL},F_{LH},F_{HL},F_{HH}\},\\
    F_{LL}=\begin{bmatrix}0.5&0.5\\0.5&0.5\end{bmatrix}, F_{LH}=\begin{bmatrix}-0.5&-0.5\\0.5&0.5\end{bmatrix},\\F_{HL}=\begin{bmatrix}-0.5&0.5\\-0.5&0.5\end{bmatrix}, F_{HH}=\begin{bmatrix}0.5&-0.5\\-0.5&0.5\end{bmatrix},
\end{aligned}
\label{eq:2}
\end{equation}
where $F_ {LL}$ is equivalent to average pooling. The WaveletPool operation can be expressed as:
\vspace{-0.2cm}
\begin{equation}
    \label{eq:3}
    Y_{{pool}}=\sum_{j=1}^4F_j*Y_{{C2f}},
\end{equation}
%\vspace{-0.1cm}
where $*$ denotes the convolution operation. This operation extracts multi-scale features from the input feature map by applying each filter $F_{j}$ and aggregating the results across the channel dimension.
\vspace{-0.1cm}
\subsubsection{Context Guided}
The ContextGuided block takes the input feature map and performs a series of operations, defined as:
\vspace{-0.1cm}
\begin{equation}
\label{eq:4}
    Y_{CGB}={CGB}(Y_{pool}).
\end{equation}
The processing steps within the ContextGuided block can be expressed as follows:
\vspace{-0.1cm}
\begin{equation}
    \begin{aligned}
    \label{eq:5}
    X_1={Conv}_{1\times1}(Y_{{pool}})\in{R}^{B\times n\times H\times W},  n=\frac{C_{{out}}}2,
    \end{aligned}
\end{equation}
where $n$ represents the number of the input channels. Then a depthwise convolution is employed to capture local features $L$, and another depthwise convolution with a dilation rate $d$ is applied to capture the surrounding features $S$:
\vspace{-0.1cm}
\begin{equation}
    \label{eq:6}
    L={Conv}_{3\times3}(X_1)\in{R}^{B\times n\times H\times W},
\end{equation}
\begin{equation}
    \label{eq:7}
    S={Conv}_{3\times3}(X_1)\quad\text{with dilation rate}\;d.
\end{equation}
The $L$ and $S$ are concatenated, normalized, and activated, generating the global context feature $F_{glo}$ through the global feature extraction module:
\begin{equation}
    \label{eq:8}
J=BatchNorm({Concat}(L,S))\in{R}^{B\times2n\times H\times W},
\end{equation}
\begin{equation}
    \label{eq:9}
G=F_\text{glo}(J)\in{R}^{B\times C_\text{out}\times H\times W}.
\end{equation}
Finally, the output of the Context Guided Block is computed by adding the input to the global context refinement output:
\vspace{-0.1cm}
\begin{equation}
    \label{eq:10}
Y_{{CGB}}=Y_{{pool}}+G.
\end{equation}
\vspace{-0.5cm}
\subsection{Waveletpool-StarFusion (WSF)}
Although the WCG module reduces parameters, this letter proposes WSF to better balance parameters and model accuracy as described in Fig. \ref{fig:fig2}. WaveletUnPool and Star module enhance SAR ship detection by restoring spatial resolution and retaining rich features for improved contour recognition. Star module utilizes depthwise separable convolutions and element-wise multiplication operations for real-time applicability while effectively mitigating noise. Together, they significantly enhance performance and efficiency in ship detection.

\subsubsection{WaveletUnPool} 
WaveletUnPool, in contrast, aims to restore the spatial resolution of the downsampled feature map, effectively reconstructing its original dimensions. It utilizes transpose convolution with a stride of 2, employing the same wavelet filters as described in formula \ref{eq:2} to upsample the feature map back to a higher resolution.
\subsubsection{Star}
%StarNet is a deep learning architecture that enhances model representation capability by utilizing Star Operation. The core idea of StarNet is to fuse features from different subspaces through element-wise multiplication at each layer, effectively expanding the model's implicit feature dimensions without significantly increasing the number of parameters. Figure \ref{fig:fig2} illustrates the process in the neck section, showing how the feature map is concatenated for the first time and then input into StarNet to obtain the corresponding feature map. It also explains the computational flow of StarNet.
Element-wise multiplication structure inside StarNet \cite{ma2024rewrite} which is an advanced deep learning architecture that enhances model representation. By fusing features from different subspaces with element-wise multiplication at each layer, it expands implicit feature dimensions without significantly increasing parameters. Fig. \ref{fig:fig2} illustrates this process in the neck section, showing the initial concatenation of the feature map before inputting it into Star module, along with the computational flow involved.

The calculation process of Star block is explained below. The Star block receives an input feature map $x_{in}$ with dimensions \(C\times H^{\prime}\times W^{\prime}\), where $C$ is the number of channels, and $H^{\prime}$ and $W^{\prime}$ are the height and width of the feature map, respectively.
\begin{itemize}
    \item First, a depthwise convolution is applied to the input feature map \(x_{in}\), and \(W_k\) is th convolution kernel for the k-th channel:
    \vspace{-0.1cm}
          \begin{equation}
          \label{eq:11}
              {x_{in}}'= DWConv(x_{in})=\sum_{k=1}^CW_k*x_{{in}},
          \end{equation}
    \item Next, $1\times1$ convolution layers are used to transform features, producing two feature maps:
           \begin{equation}
           \label{eq:12}
               \begin{aligned}
                 {x}_1'={Conv}_{1\times1}({x_{in}}')=W_{1}\cdot x_{in}'+b_{1}, \\{x}_2'={Conv}_{1\times1}({x_{in}}')=W_{2} \cdot x_{in}'+b_{2}.
               \end{aligned}
            \end{equation}
        %The number of channels in  ${x}_1'$ and ${x}_2'$ are both mlp\_ratio $\times$ $C$, with mlp\_ratio = 3.
    \item Then, the first feature map undergoes ReLU6 activation, followed by element-wise multiplication with the second feature map ${x_{in}}^{\prime\prime}$, and another another $1\times 1$ convolution layer is applied to generate a new feature map ${x}^{(new)}$:
    \vspace{-0.1cm}
    \begin{equation}
    \label{eq:13}
            {x_{in}}^{\prime\prime}=min(max(0,{x}_1^{\prime}),6)\odot{x}_2^{\prime},
    \end{equation}
    \begin{equation}
    \label{eq:14}
    \begin{aligned}
    {x}^{(new)} &= \text{DWConv}2(\text{Conv}_{1\times1}(g({x_{in}}^{\prime\prime}))) \\
    &= \sum_{k=1}^C W_{k}^{new} * g(x_{{in}}^{\prime\prime}).
    \end{aligned}
    \end{equation}

    \item Finally, a residual connection is established by adding the input feature map \({x_{in}}\) to the newly generated feature map, applying dropout for regularization:
    \begin{equation}
    \label{eq:15}
        {y}={x_{in}}+\text{dropout}({x}^{(new)})
    \end{equation}
\end{itemize}
\vspace{-0.7cm}
\subsection{LS Detector}
The FOCS \cite{lin2017focal} shows that groupnorm enhances detection head performance in localization and classification. Shared convolutions reduce model parameters, facilitating lightweight design for resource-constrained devices. To address inconsistencies in handling targets of varying scales, we introduce a scale layer for feature adjustment. The specific structure is shown in Fig. \ref{fig:fig3}. These strategies lower parameters and computational load while maintaining accuracy.
\begin{figure}
\small 
\centering
\includegraphics[width=\linewidth]{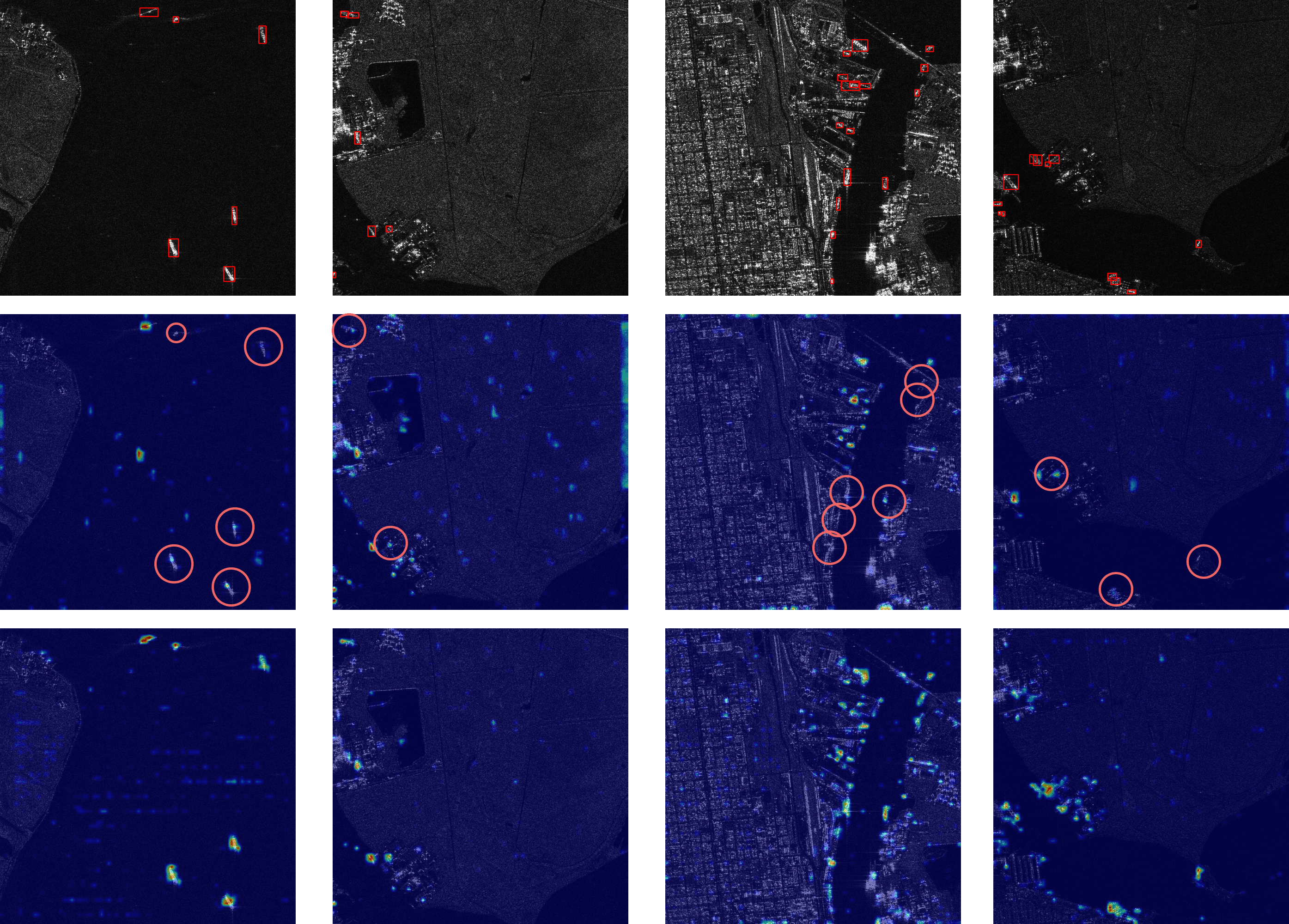}
\caption{The heat-map visualization results feature three rows: the first row illustrates the ground truth, the second row presents the baseline1 (YOLOv8n), and the third row highlights our proposed method. Red circles represent missed detections.}
\label{fig:fig4}
\medskip
\end{figure}
\begin{table}[]
    \caption{SPECIFIC EXPERIMENTAL DETAILS}
    \label{tab:table1}
    \centering
    \footnotesize % 使用较小的字体
    \setlength{\tabcolsep}{3pt} % 设置列间距为3pt，调整到你需要的值
    \begin{tabular}{ccccccc}
    \hline
         Dataset & Epochs & Optimizer & Batch & Lr0 & Momentum & Weight\_decay  \\
    \hline
         SSDD & 300 & AdamW & 16 & 0.002 & 0.937 & 5e-4 \\
         HRSID & 300 & AdamW & 16 & 0.002 & 0.937 & 5e-4 \\
    \hline
    \end{tabular}
\end{table}
\begin{figure}
\small 
\centering
\includegraphics[width=\linewidth]{fig/per_com2.png}
\caption{Performance Comparison of Three Baseline Models and the RSNet-Enhanced Model in Terms of Preprocessing Time and Inference Time.}
\label{fig:fig5}
\medskip
\end{figure}

\begin{table}[h]
    \caption{ABLATION EXPERIMENTS ON HRSID}
    \label{tab:table2}
    \centering
    \footnotesize
    \setstretch{0.9}
    \setlength{\tabcolsep}{1.5pt}
    \begin{tabular}{ccccccc}
    \hline
        &&&\multicolumn{2}{c}{Baseline1 (YOLOv8n)}&&\\
        \hline
         WCG & WSF & LS & $\mathrm{mAP_{.50}}(\%)$ & $\mathrm{mAP_{.50:.95}}(\%)$ & Params(M) & FLOPs(G)  \\
    \hline
         × & × & × & 91.1 & 66.9 & 3.01 & 8.1 \\
         \checkmark & × & × & 91.2 (0.6 $\uparrow$) & 66.5 (0.4 $\downarrow$) & 2.43 (0.58 $\downarrow$) & 7.4 (0.7 $\downarrow$) \\
         \checkmark & \checkmark & × & 91.1 & 65.8 (1.1 $\downarrow$)  & 2.13 (0.88 $\downarrow$) & 6.8 (1.3 $\downarrow$)\\
         \checkmark & \checkmark & \checkmark & \textbf{91.2} (0.1 $\uparrow$) & \textbf{67.6} (0.7 $\uparrow$) & \textbf{1.49} (1.52 $\downarrow$) & \textbf{5.1} (3.0 $\uparrow$)\\ 
    \hline
    &&&\multicolumn{2}{c}{Baseline2 (YOLOv9n)}&&\\
    \hline
    WCG & WSF & LS & $\mathrm{mAP_{.50}}(\%)$ & $\mathrm{mAP_{.50:.95}}(\%)$ & Params(M) & FLOPs(G)  \\
    \hline
         × & × & × & 90.5 & 66.1 & 2.01 & 7.8 \\
         \checkmark & × & × & 91.1 (0.6 $\uparrow$) & \textbf{66.9} (0.8 $\uparrow$) & 1.65 (0.36 $\downarrow$) & 7.2 (0.6 $\downarrow$)\\
         \checkmark & \checkmark & × & 90.3 (0.2 $\downarrow$) & 65.6 (0.5 $\downarrow$) & 1.24 (0.77 $\downarrow$) & 5.9 (1.9 $\downarrow$)\\
         \checkmark & \checkmark & \checkmark & \textbf{91.1} (0.6 $\uparrow$) & 66.6 (0.5 $\uparrow$) & \textbf{0.77} (1.24 $\downarrow$) & \textbf{4.5} (3.3 $\downarrow$)\\
    \hline
    &&&\multicolumn{2}{c}{Baseline3 (YOLOv11n)}&&\\
   \hline
        WCG & WSF & LS & $\mathrm{mAP_{.50}}(\%)$ & $\mathrm{mAP_{.50:.95}}(\%)$ & Params(M) & FLOPs(G)  \\
        \hline
        × & × & × & 89.7 & 65.8 & 2.59 & 6.4 \\
        \checkmark & × & × & 90.0 (0.3 $\uparrow$) & 66.2 (0.4 $\uparrow$) & 2.01 (0.58 $\downarrow$) & 5.2 (1.2 $\downarrow$) \\
        \checkmark & \checkmark & × & 90.0 (0.3 $\uparrow$) & 66.1 (0.3 $\uparrow$) & 1.30 (1.29 $\downarrow$) & 5.1 (1.3 $\downarrow$) \\
        \checkmark & \checkmark & \checkmark & \textbf{91.1} (0.4 $\uparrow$) & \textbf{67.0} (1.2 $\uparrow$) & \textbf{1.05} (1.54 $\downarrow$) & \textbf{4.9} (1.5 $\downarrow$) \\
        \hline
    \end{tabular}
\end{table}

%Heat-map visualization results. The first row represents the ground truth, the second row shows the baseline, and the third row illustrates our method.

\section{EXPERIMENT}

\subsection{Datasets}
%To evaluate the performance of our model, we utilized two publicly available SAR remote sensing datasets: SSDD \cite{zhang2021sar} and HRSID \cite{wei2020hrsid}. The SSDD consists of 1160 images, with 928 allocated for training and 232 for testing, as outlined in the original publication. In the case of the HRSID, we adhered to the official partitioning, employing 3642 images for training and 1962 for testing purposes.
We evaluated our model on two public SAR datasets: SSDD \cite{zhang2021sar} and HRSID \cite{wei2020hrsid}. SSDD includes 1160 images, with 928 for training and 232 for testing as per the original split. HRSID follows the official split, using 3642 training and 1962 testing images.

\subsection{Training Details}
Our experiments were conducted on a system running Ubuntu 20.04, equipped with an Intel Xeon Silver 4210R CPU and a Nvidia GeForce Tesla V100 16G. And the image size is set to 640. The specific experimental details are shown in Table \ref{tab:table1}. 
%在预处理以及推理时间上 三类基线模型与加入RSNet后的模型的性能比较图

\begin{table}[]
    \caption{COMPARISON EXPERIMENTS ON HRSID}
    \label{tab:table3}
    \centering
    \footnotesize % 使用较小的字体
    \setlength{\tabcolsep}{1.5pt} % 设置列间距
    \setstretch{0.8} % 自定义行距
    \begin{tabular}{ccccc}
    \hline
         Method &  \(\mathrm{mAP_{.50}}(\%)\) & \(\mathrm{mAP_{.50:.95}}(\%)\) & Params(M) & FLOPs(G) \\
    \hline
         Faster R-CNN-FPN & 72.0 & 46.5 & 41.35 & 134 \\
         RetinaNet \cite{lin2017focal} & 78.9 & 53.6 & 36.33 & 128 \\
         YOLOv6n \cite{li2022yolov6}& 88.2 & 62.8 & 4.23 & 11.8 \\
         YOLOv7-tiny \cite{wang2023yolov7} & 85.4 & 57.2 & 6.01 & 13.2 \\
         YOLOv8n & 91.1 & 66.9 & 3.01 & 8.1 \\
        YOLOv9t \cite{wang2024yolov9} & {90.5} & {66.1} & {2.01}&{7.8} \\
        {YOLOv10n} \cite{wang2024yolov10} & {89.6} & {66.3} & {2.71}&{8.4} \\
         YOLOv11n &89.7  & 65.8 &2.59 &6.4\\
        {YOLOv12n} \cite{tian2025yolov12}& {85.4} & {60.7} & {2.52}&{6.0} \\
    \hline
         Yue et al. \cite{yue2022generating}* & 91.1 & 66.5 & 43.42 & - \\
         CPoints-Net \cite{gao2024compact}* & 90.5 & - & 18.64 & 102.2 \\
         FEPS-Net \cite{bai2023feature}* & 90.7 & 65.7 & 37.31 & - \\
         BL-Net \cite{zhang2021balance}* & 88.67 &-&47.81&-\\
         FBUA-Net \cite{bai2023novel}* &90.3&-&36.54&-\\
         $\mathrm{{CS^{n}Net}}$ \cite{chen2023cs}* &91.2&60.6&12.2&21.7\\
    \hline
         Ours ({YOLOv8n-based})& \textbf{91.2} & \textbf{67.6} & \textbf{1.49} & \textbf{5.1} \\
    \hline
    \end{tabular}
       \begin{flushleft}
    \footnotesize{* Denotes data obtained from their papers}
    \end{flushleft}
\end{table}

\begin{table}[]
    \caption{COMPARISON EXPERIMENTS ON SSDD}
    \label{tab:table4}
    \centering
    \footnotesize % 使用较小的字体
    \setlength{\tabcolsep}{1.5pt} % 设置列间距
        \setstretch{0.8} % 自定义行距
    \begin{tabular}{ccccc}
    \hline
         Method &  \(\mathrm{mAP_{.50}}(\%)\) & \(\mathrm{mAP_{.50:.95}}(\%)\) & Params(M) & FLOPs(G)  \\
    \hline
         Faster R-CNN-FPN & 94.0 & 64.0 & 41.35 & 134 \\
         RetinaNet \cite{lin2017focal} & 90.2 & 60.0 & 36.33 & 128 \\
         YOLOv6n \cite{li2022yolov6}& 96.9 & 71.2 & 4.23 & 11.8 \\
         YOLOv7-tiny \cite{wang2023yolov7}& 96.4 & 66.5 & 6.01 & 13.2 \\
         YOLOv8n & 98.2 & \textbf{73.1} & 3.01 & 8.1 \\
         {YOLOv9t} \cite{wang2024yolov9} & {98.2} & {71.2} & {2.01}&{7.8}\\
        {YOLOv10n} \cite{wang2024yolov10} & {97.0} & {71.3} & {2.71}&{8.4} \\
         YOLOv11n & 98.0 & 72.0 &2.59 &6.4\\
        {YOLOv12n} \cite{tian2025yolov12}& {97.3} & {70.1} & {2.52}&{6.0} \\
    \hline
         Yue et al. \cite{yue2022generating}* & 95.7 & 64.8 & 43.42 & - \\
         CPoints-Net \cite{gao2024compact}* & 96.3 & - & 18.64 & 102.2 \\
         FEPS-Net \cite{bai2023feature}* & 96.0 & 73.1 & 37.31 & - \\
         BL-Net \cite{zhang2021balance}* & 95.25 &-&47.81&-\\
         FBUA-Net \cite{bai2023novel}* &96.2&-&36.54&-\\
         $\mathrm{{CS^{n}Net}}$ \cite{chen2023cs}* &97.1&64.9&12.2&21.7\\
    \hline
         Ours (YOLOv8n-based)& \textbf{98.4} & 72.5 & \textbf{1.49} & \textbf{5.1} \\
    \hline
    \end{tabular}
       \begin{flushleft}
    \footnotesize{* Denotes data obtained from their papers}
    \end{flushleft}\end{table}

\subsection{Evaluation Metrics}
%To evaluate the proposed model's performance, we used mean average precision at IoU = 0.5 (\(\mathrm{mAP_{.50}}\)) and IoU = 0.5:0.05:0.95 (\(\mathrm{mAP_{.50:.95}}\)), along with the number of parameters (Params) and floating-point operations per second (FLOPs). The mAP metrics assess detection accuracy, while Params and FLOPs indicate parameters and computational complexity.
Model performance was assessed using $\mathrm{mAP_{.50}}$ and $\mathrm{mAP_{.50:.95}}$ for detection accuracy, alongside parameter count (Params) and FLOPs to quantify model complexity and computational efficiency.

\subsection{Ablation Experiments}
To validate the module's effectiveness, we conducted ablation experiments on the HRSID dataset based on YOLOv8n, YOLOv9n and YOLOv11n, as shown in Table \ref{tab:table2}. We also performed feature visualization experiments, illustrated in Fig. \ref{fig:fig4}, which demonstrate that our model effectively identifies real targets while reducing the influence of surrounding noise. To evaluate the performance of the proposed framework under computational constraints, we test the baseline models and their RSNet-enhanced models on an Intel i5-8265U CPU (TDP = 15W) at 20°C under CPU-only inference for fair comparison. Results are shown in Fig. \ref{fig:fig5}.
%此外我们将基线模型与对应的RSNet-Enhanced Model在室温20摄氏度环境下，使用搭载i5 8265U cpu (TDP=15W) 的设备上仅使用cpu进行部署推理，以测试改进前后的模型性能，结果如图Fig. \ref{fig:fig5}.

\subsection{Comparison Experiments}
We selected several representative object detectors, including the classic two-stage Faster R-CNN-FPN \cite{ren2016faster}, one-stage RetinaNet \cite{lin2017focal}, and YOLOv6n \cite{li2022yolov6}, YOLOv7-tiny \cite{wang2023yolov7}, YOLOv8n and YOLOv11n. Additionally, we compared algorithms specifically designed for target detection in SAR images. The experimental results, presented in Tables \ref{tab:table3} and \ref{tab:table4}, demonstrate that our method effectively balances parameter quantity and performance, outperforming other methods.

\section{CONCLUSIONs}
This letter introduces RSNet, a lightweight network for SAR ship detection that enhances accuracy and efficiency through its innovative architecture. The WCG backbone improves performance with fewer parameters, facilitating effective feature extraction and context understanding. The WSF neck optimizes feature integration, while the LS module ensures efficient detection with minimal computational overhead. Future work will explore RSNet's application to other detection tasks and datasets, further expanding its utility in remote sensing.

\bibliographystyle{IEEEtran}
\bibliography{references}

\vfill
\end{document}